\documentclass{article}

\usepackage[nonatbib,preprint]{neurips_2026}

\usepackage[utf8]{inputenc} 
\usepackage[T1]{fontenc}    
\usepackage{url}            
\usepackage{booktabs}       
\usepackage{amsmath, amsfonts}       
\usepackage{nicefrac}       
\usepackage{microtype}      
\usepackage{xcolor}         
\usepackage{graphicx}
\usepackage{multirow}
\usepackage{subfig}
\usepackage{wrapfig}
\usepackage{enumitem}
\usepackage[linesnumbered, ruled]{algorithm2e}
\usepackage[colorlinks,
            linkcolor=blue,
            anchorcolor=blue,
            citecolor=blue]{hyperref}
\usepackage{utfsym} 
\newcommand{\heart}{$\;\!$\usym{2665}}

\newcommand{\projectname}{ICED}
\newcommand{\partitle}[1]{\noindent \textbf{#1.}}

\usepackage{amsthm}
\newtheorem{theorem}{Theorem}
\newtheorem{remark}{Remark}

\title{ICED: Concept-level Machine Unlearning via Interpretable Concept Decomposition}

\author{
Shen Lin$^{1}$, Jing Lin$^{1}$, Junhao Dong$^{2}$\thanks{Corresponding author.}\ \ , Piotr Koniusz$^{3,4}$, and Li Xu$^{1}$\\ [2mm]
$^{1}$Fujian Normal University, 
$^{2}$Nanyang Technological University, \\[1mm]
$^{3}$University of New South Wales, 
$^{4}$Data61$\!${\color{red}\heart}CSIRO \\[1mm]
}

\begin{document}

\maketitle

\begin{abstract}
Machine unlearning in Vision-Language Models (VLMs) is typically performed at the image or instance level, making it difficult to precisely remove target knowledge without affecting unrelated semantics. This issue is especially pronounced since a single image often contains multiple entangled concepts, including both target concepts to be forgotten and contextual information that should be preserved. In this paper, we propose an interpretable concept-level unlearning framework for VLMs, which constructs a compact task-specific concept vocabulary from the forgetting set using a multimodal large language model. In addition to modality alignment, visual representations are decomposed into sparse, nonnegative combinations of semantic concepts, providing an explicit interface for fine-grained knowledge manipulation. Based on this decomposition, our method formulates unlearning as concept-level optimization, where target concepts are selectively suppressed while intra-instance non-target semantics and global cross-modal knowledge are preserved. Extensive experiments across both in-domain and out-of-domain forgetting settings demonstrate that our method enables more comprehensive target forgetting, better preserves non-target knowledge within the same image, and maintains competitive model utility compared with existing VLM unlearning methods.
\end{abstract}

\section{Introduction}
Recent advances in large-scale Vision–Language Models (VLMs), exemplified by CLIP \cite{radford2021learning}, have established a powerful framework for learning aligned representations. Despite these successes, the practical deployment of such models has raised growing concerns about data governance, privacy, robustness, and regulatory compliance \cite{li2023trustworthy, dong2026allies, dongtug, lin2025deepaw}, as they may implicitly retain and expose sensitive, copyrighted, or harmful information present in their training set. This challenge is further amplified by recent regulatory trends that emphasize the right to data removal and accountability in AI systems. Machine unlearning \cite{chundawat2023can,lin2023erm,chen2023boundary,lin2024gdr,patel2025learning} has thus emerged for selectively erasing the influence of specific training data from a trained model without requiring costly retraining from scratch.

Existing VLM unlearning methods \cite{poppi2024safe,cheng2024multidelete,cai2025targeted,yang2025cliperase,zhang2025targeted,dong2025machine} often perform forgetting at the image or sample level, treating each training instance as an atomic unit for removal. Most methods operate at the parameter or embedding level, treating unlearning as an optimization problem over model weights without explicitly modeling the semantic structure of vision--language alignment. This lack of semantic grounding makes it difficult to localize the exact knowledge being forgotten, often leading to the entangled or incomplete removal of target concepts.

\begin{figure}[htbp]
    \centering
    \includegraphics[width=0.9\linewidth]{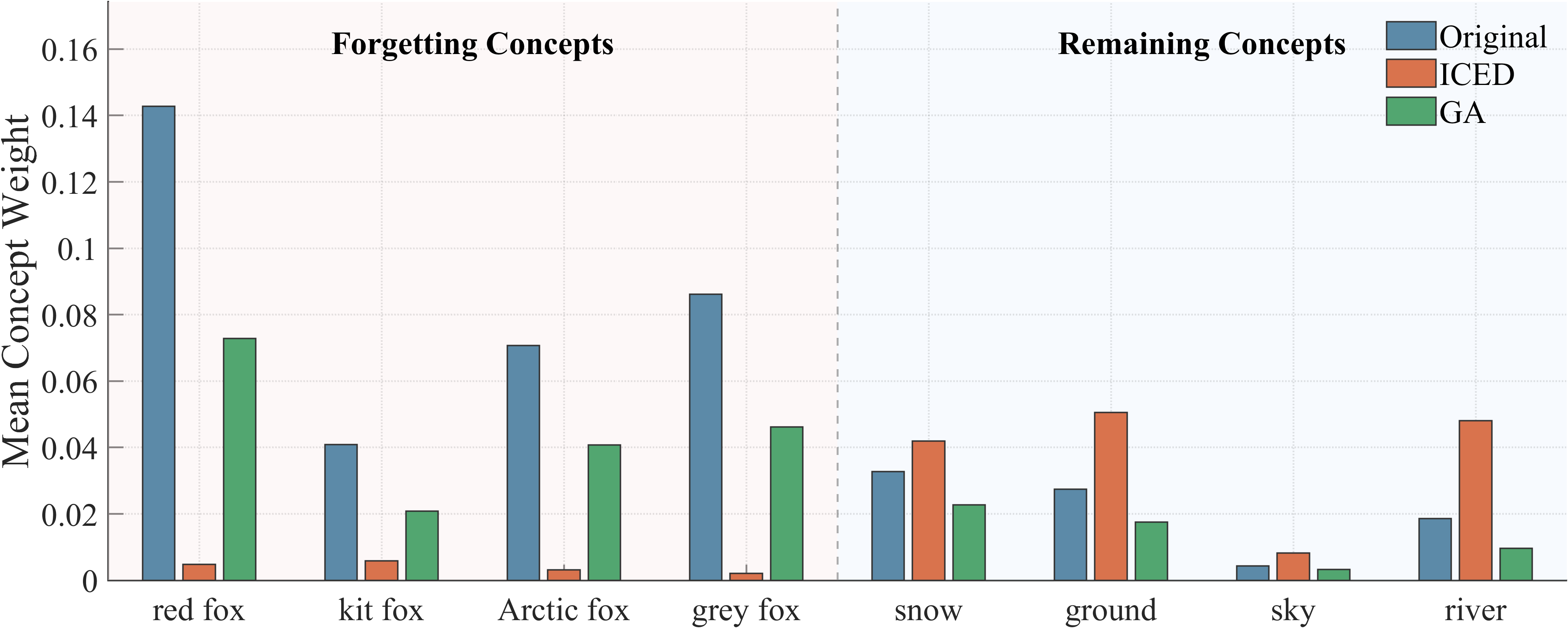}
    \caption{\textbf{Motivation.} The left part shows target concepts to be forgotten, while the right part shows remaining contextual concepts that should be preserved. ICED more effectively suppresses forgetting concepts and shifts the model's focus toward non-target contextual concepts, indicating more selective and utility-preserving unlearning.}
    \label{fig:concept_difference}
    \vspace{-1em}
\end{figure}

Moreover, this limitation becomes more evident when accounting for the internal structure of visual inputs. A single image often contains multiple semantic concepts, including both target concepts intended for forgetting and irrelevant contextual information that should be preserved. However, existing methods typically treat each instance as an atomic unit for removal. As illustrated in Fig.~\ref{fig:concept_difference}, conventional unlearning methods can reduce the weights associated with target concepts to some extent, but they may also suppress or distort non-target concepts. Once the target concepts are effectively removed, the model should instead redirect its attention to the remaining non-target concepts, rather than uniformly weakening the entire representation. In contrast, our method more selectively suppresses the forgetting concepts while preserving stronger activations on the surrounding contextual concepts, indicating a more fine-grained concept-level forgetting behavior. Consequently, removing an entire instance may inadvertently eliminate unrelated yet useful knowledge, resulting in unnecessary degradation of general vision--language capabilities. These observations suggest that image-level unlearning is too coarse to disentangle target concepts from co-occurring non-target semantics, making it difficult to achieve precise forgetting without compromising model utility.

This motivates the need for fine-grained, concept-level unlearning in VLMs, where semantic components within each instance can be explicitly identified and manipulated. In this work, we propose a novel machine unlearning method based on \textbf{I}nterpretable \textbf{C}onc\textbf{E}pt \textbf{D}ecomposition, dubbed \projectname. Specifically, \projectname \ decomposes CLIP representations into sparse, non-negative, and overcomplete concept bases, enabling fine-grained localization of semantic concepts within individual instances. This structured decomposition supports the selective removal of target concepts while preserving irrelevant but useful visual-language knowledge. We summarize our contributions as follows:

\begin{enumerate}[label=\Roman*)]
    \item We identify a key limitation of existing vision-language model unlearning methods: instance-level forgetting induces entangled representations and non-selective knowledge removal. To address this limitation, we propose ICED, an interpretable concept-level unlearning framework that integrates concept decomposition with concept-level optimization.

    \item We develop ICED with two key components: an interpretable concept decomposition module that constructs a task-specific concept vocabulary and represents each image as a combination of semantic concepts, and a concept-level optimization objective that removes target concepts while preserving non-target semantics and global cross-modal knowledge.

    \item We conduct extensive experiments on CIFAR-10 and ImageNet-1K with two CLIP backbones, comparing ICED against state-of-the-art baselines. The results show that ICED achieves a superior trade-off between effective unlearning and model preservation.
\end{enumerate}

\section{Related Work}

\partitle{CLIP Interpretability} 
Recent works leverage the semantic structure of CLIP and its text encoder to interpret visual representations through concept-based analysis. A common line of research represents image embeddings via their similarities to textual concepts, enabling downstream concept bottleneck models or probing methods~\cite{moayeri2023text,yuksekgonul2023posthoc,yun2023do}. 
Another direction focuses on structured concept decompositions, expressing image representations as combinations of human-interpretable concepts through sparse coding, latent dictionaries, or attribution mechanisms~\cite{chen2023stair,gandelsman2024interpreting,chattopadhyay2023information,bhalla2024interpreting, dong2025robust}. Despite these advances, existing CLIP interpretability methods are not well-suited for machine unlearning. They primarily provide post-hoc explanations rather than controllable representations. Moreover, they lack mechanisms for selectively suppressing identified concepts while preserving model knowledge.

\partitle{Machine unlearning for CLIP} 
Existing works \cite{poppi2024safe,cheng2024multidelete,cai2025targeted, yang2025cliperase, zhang2025targeted, dong2025machine} have explored machine unlearning in CLIP models by targeting the removal of specific concepts or classes. For example, Cheng et al. \cite{cheng2024multidelete} proposed a multi-modal unlearning method, namely MultiDelete, which is designed to separate embeddings for the forget set while preserving uni-modal embeddings for the retain set. Cai et al. \cite{cai2025targeted} computed unlearning gradients on the forgetting data and restricted the update to a single selected layer. Yang et al. \cite{yang2025cliperase} proposed CLIPErase, which disentangles and selectively removes both visual and textual associations in CLIP. Kravets et al. \cite{kravets2025zero} employed Lipschitz regularization with synthetic samples for zero-shot class unlearning. 

Different from existing works, we introduce an interpretable concept decomposition of CLIP representations, enabling direct intervention on semantically meaningful components and supporting precise, interpretable concept-level machine unlearning.

\section{Proposed Method}

\subsection{Overview}
\begin{figure}[htbp]
    \centering
    \includegraphics[width=1\linewidth]{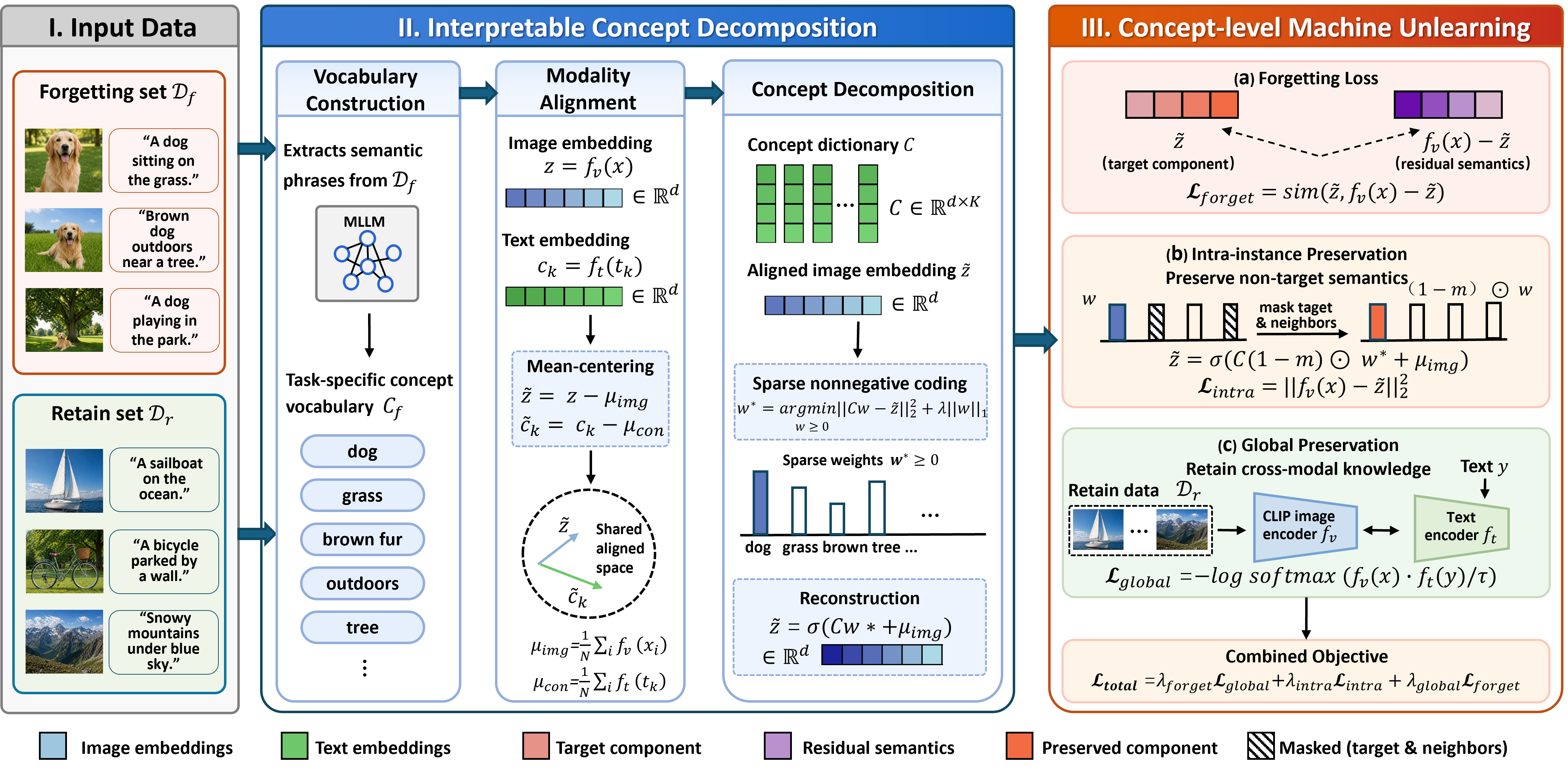}
    \caption{An overview of our proposed \projectname \ method.}
    \label{fig:overview}
    \vspace{-1em}
\end{figure}

As shown in Fig.~\ref{fig:overview}, we present an interpretable framework for concept-level machine unlearning in vision-language models. Instead of removing entire instances, our method decomposes visual representations into sparse combinations of semantic concepts and selectively suppresses the target ones.
The framework consists of two stages. First, we construct a task-specific concept vocabulary from the forgetting set using an MLLM, and align it with the visual embedding space. Each image is then represented as a sparse, nonnegative combination of concept embeddings, yielding an interpretable decomposition.
Second, we perform concept-level unlearning by directly manipulating these representations. We design three objectives: a forgetting loss to remove target concepts from remaining semantics, an intra-instance preservation loss to maintain non-target information within each sample, and a global preservation loss to retain general knowledge on the remaining data.
This formulation enables precise and controllable unlearning while preserving overall model performance.

\subsection{Interpretable Concept Decomposition}

\partitle{Vocabulary Construction}
Unlike prior work that relies on a predefined universal concept set, we construct a task-specific concept vocabulary directly from the forgetting set. A unified large-scale vocabulary typically introduces two issues in our setting: it increases storage overhead due to the need to maintain a large concept dictionary, and it may degrade representation precision because many irrelevant or redundant concepts can interfere with the decomposition process.

Given a forgetting set $\mathcal{D}_f$, we leverage a multimodal large language model (MLLM) to extract a set of semantic concepts $\mathcal{C}_f = \{c_k\}_{k=1}^{K}$ that capture the dominant objects, attributes, and contextual patterns present in $\mathcal{D}_f$. Specifically, for each sample $x \in \mathcal{D}_f$, we prompt the MLLM with both visual and textual inputs to generate descriptive phrases grounded in the image content, and aggregate the resulting concepts across the dataset to form an initial candidate pool:
\begin{equation}
    \mathcal{C}_f = \bigcup_{x \in \mathcal{D}_f} \text{MLLM}(x).
\end{equation}
To improve robustness and reduce noise, we further perform deduplication and frequency-based filtering, retaining only the most representative and semantically consistent concepts. We also supplement the vocabulary with a small set of common background concepts and their synonyms to improve the coverage of non-target contextual semantics. Compared to generic concept vocabularies, this data-driven construction yields a compact and task-relevant concept set that is more tightly aligned with the forgetting objective.

\paragraph{Modality Alignment.}
To decompose visual features into nonnegative combinations of semantic concepts, the image embeddings and concept embeddings should lie in a shared and compatible space. However, due to the modality gap \cite{dong2025stabilizing} in CLIP, the two modalities may occupy different regions on the unit sphere, which hinders reliable nonnegative decomposition. To address this issue, we perform mean-centering separately for image embeddings and concept embeddings.
Let $\mathbf{z}_{\text{img}} = f_v(x)$ denote the CLIP image embedding and $g(c_k)=f_t(t_k)$ denote the text embedding of concept $c_k$. We estimate the image-space mean $\boldsymbol{\mu}_{\text{img}}$ and the concept-space mean $\boldsymbol{\mu}_{\text{con}}$, and center the embeddings as
\begin{equation}
    \tilde{\mathbf{z}} = \sigma(\mathbf{z}_{\text{img}} - \boldsymbol{\mu}_{\text{img}}), \quad
    \tilde{\mathbf{c}}_k = \sigma(g(c_k) - \boldsymbol{\mu}_{\text{con}}),
\end{equation}
where $\sigma(\cdot)$ denotes $\ell_2$ normalization. This alignment alleviates the modality gap and enables reliable nonnegative decomposition in the shared space. After obtaining the reconstructed embedding $\hat{z}$, we map it back to the original CLIP image space as
\begin{equation}
\hat{\mathbf{z}}_{\text{img}} = \sigma(\hat{\mathbf{z}} + \boldsymbol{\mu}_{\text{img}}).
\end{equation}
This ensures that the reconstructed representation lies on the same cone as the original CLIP image embeddings, which is important for preserving downstream performance after concept-level manipulation.

\paragraph{Concept Decomposition.}
Given the aligned concept vocabulary $\mathcal{C} = \{c_k\}_{k=1}^{K}$ constructed from the forgetting set and the centered CLIP image embedding $\mathbf{z} = \sigma(\mathbf{z}_{\text{img}} - \boldsymbol{\mu}_{\text{img}})$, we aim to represent $z$ as a sparse nonnegative combination of concept embeddings. Let $C = [\sigma(g(c_1) - \boldsymbol{\mu}_{\text{con}}), \cdots, \sigma(g(c_K) - \boldsymbol{\mu}_{\text{con}})]$ denote the centered and normalized concept dictionary, where $g(\cdot)$ is the CLIP text encoder and $\sigma(\cdot)$ denotes $\ell_2$ normalization. We formulate concept decomposition as an optimization problem:
\begin{equation}
\min_{\boldsymbol{w} \in \mathbb{R}_+^K} \|\boldsymbol{w}\|_0 \quad \text{s.t.} \quad \langle \mathbf{z}, \sigma(C\boldsymbol{w}) \rangle \geq 1 - \epsilon,
\end{equation}
where $\epsilon$ controls the reconstruction tolerance. Following standard practice, we relax the $\ell_0$ constraint by introducing an $\ell_1$ sparsity regularization, yielding the following convex surrogate:
\begin{equation}
\boldsymbol{w}^*=\arg\min_{\boldsymbol{w} \in \mathbb{R}_+^K} \|C\boldsymbol{w} - \mathbf{z}\|_2^2 + \lambda_{dec} \|\boldsymbol{w}\|_1,
\end{equation}
where $\lambda_{dec}$ controls the strength of the sparsity regularization.

\subsection{Concept-level Machine Unlearning}

Based on the proposed interpretable concept decomposition, we formulate machine unlearning as a concept-level optimization problem. Instead of removing entire instances, our goal is to selectively suppress target concepts while preserving both unrelated contextual information within the same instance and general knowledge learned from the remaining data.

\partitle{Forgetting Loss}
To remove target concepts while avoiding interference with other unrelated semantics within the same image, we enforce a separation between the target concept embedding and the remaining concept representations. Specifically, we minimize the similarity between the reconstructed embedding of the forgetting concept and the residual visual representation that captures non-target concepts:
\begin{equation}
\mathcal{L}_{\text{forget}} = 
\mathbb{E}_{x \sim \mathcal{D}_f} 
\left[ \mathrm{sim} \big( \mathbf{\hat{z}}, f_v(x) - \mathbf{\hat{z}} \big) \right],
\end{equation}
where $\mathrm{sim}(\cdot,\cdot)$ denotes the cosine similarity, $f_v(x)$ is the visual embedding, and $\mathbf{\hat{z}}=\sigma(C\boldsymbol{w}^\ast + \boldsymbol{\mu}_{\text{img}})$ represents the reconstructed embedding. This objective encourages the model to decorrelate the target concept from the remaining non-target semantic components within the same instance, thereby achieving more precise and localized forgetting.

\partitle{Intra-instance Preservation Loss}
To preserve non-target semantic information within the same instance, we reconstruct the visual representation after removing the influence of the forgetting concept and its semantic neighbors. Specifically, we first construct a binary mask $\boldsymbol{m}$ over the concept weights $\boldsymbol{w}$, where entries corresponding to the target concept and its synonyms are set to zero. The remaining concept weights are then used to reconstruct the embedding as follows:
\begin{equation}
\mathbf{\tilde{z}} = \sigma \big( C \big( (1 - \boldsymbol{m}) \odot \boldsymbol{w} \big) + \boldsymbol{\mu}_{\text{img}} \big),
\end{equation}
where $w$ denotes the concept weight vector, $C$ is the concept basis, $\boldsymbol{\mu}_{\text{img}}$ is the image bias term, $\odot$ denotes element-wise multiplication, and $\sigma(\cdot)$ is the activation function. We then enforce consistency between the original visual representation and the reconstructed embedding via a mean squared error objective:
\begin{equation}
\mathcal{L}_{\text{intra}} = 
\mathbb{E}_{x \sim \mathcal{D}_f} 
\left[ \left\| f_v(x) - \mathbf{\tilde{z}} \right\|_2^2 \right].
\end{equation}
This objective encourages the model to preserve the remaining semantic structure within each instance while explicitly excluding the influence of the forgetting concept and its related semantic components, thereby preventing unintended distortion of non-target knowledge during unlearning.

\partitle{Global Preservation Loss}
To preserve general semantic knowledge during unlearning, we impose a global alignment constraint on the retained dataset $\mathcal{D}_r$. Specifically, we adopt a contrastive objective consistent with CLIP, while only optimizing the image encoder to maintain stable visual representations. Formally, the global preservation loss is defined as:
\begin{equation}
\mathcal{L}_{\text{global}} = 
- \mathbb{E}_{(x,y)\sim\mathcal{D}_r}
\left[
\log \mathrm{softmax}(f_v(x)\cdot f_t(y)/\tau)
\right],
\end{equation}
where $\mathcal{D}_r$ denotes the retained dataset and $\tau$ is a temperature parameter. This objective preserves the global cross-modal alignment learned from the retained data, ensuring that general semantic knowledge remains intact while selectively removing the target concepts.

\partitle{Overall Loss}
The final objective integrates the three complementary components as follows:
\begin{equation}
\mathcal{L}_{\text{total}} = 
\lambda_{\text{forget}} \mathcal{L}_{\text{forget}} 
+ \lambda_{\text{intra}} \mathcal{L}_{\text{intra}} 
+ \lambda_{\text{global}} \mathcal{L}_{\text{global}},
\end{equation}
where $\lambda_{\text{forget}}$, $\lambda_{\text{intra}}$, and $\lambda_{\text{global}}$ are hyperparameters that balance the trade-off between effective forgetting and knowledge preservation.

\section{Theoretical Analysis}
\label{sec:theory}

\begin{theorem}[Selectivity of concept-level unlearning]
\label{thm:selective_unlearning}
Let $C=[C_T,C_R]\in\mathbb{R}^{d\times K}$ denote the aligned concept dictionary, where $C_T$ contains the target concepts to be forgotten, and $C_R$ contains the remaining non-target concepts. Assume that all concept atoms are $\ell_2$-normalized. For an image $x$, suppose its centered visual representation admits the decomposition
\begin{equation}
h(x) = C_T \boldsymbol{w}_T + C_R \boldsymbol{w}_R + r,
\qquad \boldsymbol{w}_T,\boldsymbol{w}_R\geq 0,\qquad \|\boldsymbol{r}\|_2\leq \epsilon_{\rm dec},
\end{equation}
where $w_T$ and $w_R$ are the target and non-target concept coefficients, and $r$ is the decomposition residual. Let the concept-erased representation be
\begin{equation}
\tilde h(x)=C_R \boldsymbol{w}_R+\boldsymbol{r}.
\end{equation}
Consider a unit-norm target text query $\boldsymbol{p}_T$ and a unit-norm non-target query $\boldsymbol{p}_R$. Suppose there exist constants $\alpha,\beta,\eta\geq 0$ such that
\begin{equation}
\langle \boldsymbol{p}_T,\boldsymbol{c}_i\rangle \geq \alpha \quad \forall \boldsymbol{c}_i\in C_T,\qquad
|\langle \boldsymbol{p}_T,\boldsymbol{c}_j\rangle| \leq \beta \quad \forall \boldsymbol{c}_j\in C_R,
\end{equation}
and
\begin{equation}
|\langle \boldsymbol{p}_R,\boldsymbol{c}_i\rangle|\leq \eta \quad \forall \boldsymbol{c}_i\in C_T .
\end{equation}
Then the erased representation satisfies
\begin{align}
\langle \boldsymbol{p}_T,h(x)\rangle-\langle \boldsymbol{p}_T,\tilde h(x)\rangle
&\geq \alpha \|\boldsymbol{w}_T\|_1, \label{eq:target_drop}\\
|\langle \boldsymbol{p}_R,h(x)\rangle-\langle \boldsymbol{p}_R,\tilde h(x)\rangle|
&\leq \eta \|\boldsymbol{w}_T\|_1, \label{eq:retain_change}\\
|\langle \boldsymbol{p}_T,\tilde h(x)\rangle|
&\leq \beta \|\boldsymbol{w}_R\|_1+\epsilon_{\rm dec}. 
\label{eq:target_leakage}
\end{align}
Consequently, when target concepts are well aligned with $p_T$ and weakly aligned with non-target queries, i.e., $\alpha\gg \eta$, suppressing $C_Tw_T$ produces a large decrease in the target score while causing only a small change to non-target semantic scores.
\begin{proof}
Proof can be found in Appendix \ref{sec:proof}.
\end{proof}
\end{theorem}

\begin{remark}
The theorem gives a simple explanation for why concept-level unlearning can be selective. Removing the target concept coefficients reduces the target score, while the effect on unrelated concepts depends only on how much those concepts overlap with the removed target concepts. Thus, when the concept dictionary is well separated, ICED can forget the specified concepts without substantially changing other visual semantics. This also suggests that using a clean and compact concept vocabulary is important: highly overlapping concepts make selective unlearning harder. 
\end{remark}

\section{Experiments}

\label{sec:experiments}
\subsection{Experimental Setup} 
\partitle{Datasets and Models}
Following previous works \cite{zhang2025targeted}, we evaluate \projectname\ under two settings: in-domain forgetting and out-of-domain forgetting, using two CLIP backbones, RN50 and RN101. For in-domain forgetting, we use Breeds \cite{santurkar2020breeds}, a subset of ImageNet-1K \cite{deng2009imagenet} that is closer to CLIP's pre-training distribution. For each of its 17 super-classes, we select one subgroup as the forgetting set and use the remaining three subgroups as the retaining set. For out-of-domain forgetting, we use CIFAR-10 \cite{krizhevsky2009learning}, whose distribution differs substantially from CLIP's training data.

\partitle{Implementation Details}
During unlearning, we fine-tune the CLIP model for 5 epochs using AdamW with a batch size of 192, a weight decay of 0.1, and gradient clipping with a maximum norm of 1.0. The learning rate is set to the order of $1\times10^{-6}$ for CIFAR-10 and $1\times10^{-7}$ for ImageNet. All images are resized to $224\times224$ and normalized following the standard CLIP preprocessing protocol. For vocabulary construction, we use ChatGPT to extract candidate semantic concepts from the forgetting set. Unless otherwise specified, we set the sparsity regularization weight to $\lambda_{\text{dec}}=0.35$ and the balancing coefficients to $\lambda_{\text{forget}}=0.5$, $\lambda_{\text{intra}}=95$, and $\lambda_{\text{global}}=0.075$.

\partitle{Baseline Methods} We compare \projectname~ with the following baselines: FT \cite{warnecke2023machine}, GA \cite{thudi2022unrolling}, Fisher \cite{golatkar2020eternal}, LIP \cite{foster2024information}, EMMN \cite{chundawat2023zero}, CLIP-LIP \cite{kravets2025zero}, and TIFS \cite{zhang2025targeted}. For fair comparison, all baselines are evaluated under the same forgetting and retaining splits with the same CLIP backbones. We provide detailed experimental settings for all baselines in the supplementary material.

\partitle{Metrics} Following prior work \cite{zhang2025targeted}, we report zero-shot classification accuracy on the target set (\textit{Target}), retained set (\textit{Retain}), and full dataset (\textit{All}). To evaluate model fidelity, we further test on several unseen benchmarks, including Food \cite{bossard2014food}, STL \cite{coates2011analysis}, and ObjectNet \cite{barbu2019objectnet}. We use a normalized score $\min(Acc_{\text{unlearn}} / Acc_{\text{original}}, 1)$, capped at 1 to indicate no performance degradation relative to the original model. The \textit{Avg. Score} is computed by averaging normalized scores across all evaluated datasets, serving as an overall measure of the forgetting--preservation trade-off. All results are averaged over three independent runs and the best results are highlighted in \textbf{bold}.

\subsection{Main Results}

\partitle{Comparison Experiments in In-domain Forgetting}
We first evaluate \projectname\ under the in-domain forgetting setting on ImageNet-1K. As shown in Table~\ref{tab:imagenet_1}, existing methods generally exhibit a clear trade-off between forgetting effectiveness and model preservation. FT, GA, Fisher, and LIP can reduce the target accuracy, but they substantially degrade retained-class accuracy, overall accuracy, and transfer performance on unseen datasets, indicating that their forgetting behavior is not sufficiently selective. EMMN and CLIP-LIP better preserve the original model utility in some cases, but they often leave high target accuracy, suggesting incomplete removal of the target concepts. TIFS achieves stronger forgetting than most baselines, yet still incurs noticeable utility loss. In contrast, \projectname\ consistently achieves effective target forgetting while maintaining strong retained, overall, and out-of-domain generalization performance across different CLIP backbones. 

\begin{table*}[htbp]
\centering
\caption{Performance comparison with several baselines in \textit{box-turtle} forgetting on ImageNet-1K. Results are reported as $a_b$, where $a$ is the zero-shot accuracy (\%) and $b$ is the normalized score. Additional results are provided in the supplementary material.}
\resizebox{\textwidth}{!}{
\begin{tabular}{ccccccccccc}
\toprule
\multirow{2}{*}{Backbone} & \multirow{2}{*}{Method} & \multicolumn{3}{c}{ImageNet} & \multirow{2}{*}{Food$\uparrow$} & \multirow{2}{*}{STL$\uparrow$} & \multirow{2}{*}{ObjectNet$\uparrow$} & \multirow{2}{*}{CIFAR-10$\uparrow$} & \multirow{2}{*}{Avg. Score$\uparrow$} \\ \cmidrule{3-5}
& & Target$\downarrow$ & Retain$\uparrow$ & All$\uparrow$ & & & &  &  \\ 

\midrule

\multirow{9}{*}{RN50}
& Original  
            & $68.00$ & $48.67$ & $51.94$ & $76.49$ 
            & $93.75$ & $25.83$ & $68.84$ & $-$\\ \cmidrule{2-10}
& FT  \cite{warnecke2023machine}        
            & $4.00_{5.88}$  & $0.26_{0.53}$  & $24.80_{47.75}$ & $30.34_{39.67}$ & $67.05_{71.52}$ & $9.58_{37.09}$  & $12.26_{17.81}$ & $44.07$ \\
& GA \cite{thudi2022unrolling}       
            & $14.50_{21.32}$ & $35.40_{72.73}$ & $35.39_{68.14}$ & $51.39_{67.18}$ & $77.76_{82.94}$ & $14.18_{54.90}$ & $14.08_{20.45}$ & $63.58$ \\
& Fisher \cite{golatkar2020eternal}    
            & $0.20_{0.29}$  & $0.34_{0.70}$  & $1.09_{2.10}$  & $0.23_{0.30}$  & $10.69_{11.40}$ & $1.14_{4.41}$  & $10.12_{14.70}$ & $19.04$\\
& LIP \cite{foster2024information}      
            & $0.95_{1.40}$  & $1.75_{3.60}$  & $1.14_{2.19}$  & $0.23_{0.30}$  & $10.50_{11.20}$ & $0.59_{2.28}$  & $10.46_{15.19}$ & $19.06$\\
& EMMN \cite{chundawat2023zero}     
            & $10.00_{14.71}$ & $36.67_{75.34}$ & $32.25_{62.09}$ & $47.48_{62.07}$ & $65.21_{69.56}$ & $9.65_{37.36}$  & $11.75_{17.07}$ & $58.40$ \\
& CLIP-LIP \cite{kravets2025zero} 
            & $37.00_{54.41}$ & $52.77_{100.00}$ & $53.78_{100.00}$ & $76.47_{99.97}$ & $93.80_{100.00}$ & $25.80_{99.88}$ & $67.77_{98.45}$ & $91.98$ \\
& TIFS \cite{zhang2025targeted}    
            & $0.00_{0.00}$  & $44.67_{91.78}$ & $47.34_{91.14}$ & $79.00_{100.00}$ & $83.05_{88.59}$ & $16.95_{65.62}$ & $83.80_{100.00}$ & $91.01$ \\ \cmidrule{2-10}
& \projectname \ (ours)        
            & $1.20_{1.76}$ & $50.68_{100.00}$ & $50.43_{97.09}$ & $71.66_{93.69}$ & $91.35_{97.44}$ & $20.91_{80.95}$ & $65.06_{94.51}$ & $\textbf{94.56}$ \\

\midrule
\multirow{8}{*}{RN101}
& Original  
            & $80.00$ & $54.67$ & $54.33$ & $81.16$ & $96.46$ & $29.16$ & $73.82$ & $-$\\ \cmidrule{2-10}
& FT  \cite{warnecke2023machine}        
            & ${0.00}_{0.00}$ & $0.51_{0.93}$  & $27.62_{50.84}$ & $38.54_{47.49}$ & $74.75_{77.49}$ & $13.26_{45.47}$ & $15.78_{21.38}$ & $49.09$\\
& GA  \cite{thudi2022unrolling}      
            & $32.00_{40.00}$ & $38.76_{70.90}$ & $38.75_{71.32}$ & $58.44_{72.01}$ & $84.45_{87.55}$ & $18.62_{63.85}$ & $18.73_{25.37}$ & $64.43$\\
& Fisher \cite{golatkar2020eternal}    
            & $1.12_{1.40}$  & $0.38_{0.70}$  & $0.27_{0.50}$  & $0.24_{0.30}$  & $9.94_{10.30}$  & $0.41_{1.41}$  & $13.44_{18.21}$ & $18.57$\\
& LIP  \cite{foster2024information}     
            & $0.24_{0.30}$  & $0.27_{0.49}$  & $1.74_{3.20}$  & $0.16_{0.20}$  & $11.19_{11.60}$ & $0.79_{2.71}$  & $11.89_{16.11}$ & $19.14$ \\
& EMMN \cite{chundawat2023zero}    
            & $68.00_{85.00}$ & $29.33_{53.65}$ & $26.43_{48.65}$ & $42.44_{52.29}$ & $72.46_{75.12}$ & $11.97_{41.05}$ & $17.31_{23.45}$ & $44.17$\\
& CLIP-LIP \cite{kravets2025zero} & 
            $2.00_{2.50}$  & $50.84_{92.99}$ & $51.49_{94.77}$ & $78.50_{96.72}$ & $96.42_{99.96}$ & $27.66_{94.86}$ & $70.08_{94.93}$ & $95.96$ \\ 
& TIFS \cite{zhang2025targeted} 
            & $0.00_{0.00}$ & $50.17_{91.77}$ & $49.51_{91.13}$ & $70.50_{86.87}$ & $85.45_{88.59}$ & $19.13_{65.60}$ & $89.86_{100.00}$ & $89.14$\\ \cmidrule{2-10}
& \projectname \ (ours)
            & $2.00_{2.50}$ & $53.67_{98.17}$ & $53.42_{98.33}$ & $78.57_{96.81}$ & $95.92_{99.44}$ & $26.86_{92.11}$ & $70.02_{94.85}$ & $\textbf{96.74}$ \\

\bottomrule
\end{tabular}}
\label{tab:imagenet_1}
\end{table*}

\partitle{Comparison Experiments in Out-of-domain Forgetting}
We further evaluate \projectname\ on CIFAR-10, which represents the out-of-domain forgetting setting. As reported in Table~\ref{tab:cifar10_1}, the same trade-off can be observed for existing methods. FT and GA reduce the accuracy of the target class but also impair the retained classes and overall performance. Fisher and LIP drive the target accuracy close to zero in several cases, but this comes with severe degradation of model utility. Although EMMN and CLIP-LIP preserve part of the original model behavior, their forgetting performance is unstable or insufficient. Compared with these baselines, \projectname\ achieves a better balance between target removal and knowledge preservation, maintaining stronger retained accuracy, overall accuracy, and transfer performance. These results demonstrate that concept-level unlearning enables more precise removal of target knowledge with less collateral damage to non-target semantics and general knowledge.

\begin{table*}[htbp]
\centering
\caption{Performance comparison with several baselines in \textit{Airplane} forgetting on CIFAR-10. Additional results are provided in the supplementary material.}
\resizebox{\textwidth}{!}{
\begin{tabular}{cccccccccc}
\toprule
\multirow{2}{*}{Backbone} & \multirow{2}{*}{Method} & \multicolumn{3}{c}{CIFAR-10} & \multirow{2}{*}{Food$\uparrow$} & \multirow{2}{*}{STL$\uparrow$} & \multirow{2}{*}{ObjectNet$\uparrow$} & \multirow{2}{*}{ImageNet$\uparrow$} & \multirow{2}{*}{Avg. Score$\uparrow$} \\ 
\cmidrule{3-5}
& & Target$\downarrow$ & Retain$\uparrow$ & All$\uparrow$ & & & & & \\ 

\midrule

\multirow{9}{*}{RN50}
& Original
            & $54.10$ & $66.61$ & $65.37$ & $76.49$ & $93.75$ & $25.83$ & $53.81$ &$-$\\ 
\cmidrule{2-10}
& FT  \cite{warnecke2023machine}
            & ${22.40}_{41.40}$ & $63.29_{95.02}$ & $58.20_{89.03}$ & $1.70_{2.22}$ & $45.20_{48.21}$ & $0.29_{1.12}$ & $0.00_{0.00}$ & $42.03$\\
& GA  \cite{thudi2022unrolling}
            & $5.90_{10.91}$ & $27.01_{40.55}$ & $24.90_{38.09}$ & $1.82_{2.38}$ & $34.16_{36.44}$ & $0.88_{3.41}$ & $0.40_{0.74}$ & $30.10$\\
& Fisher \cite{golatkar2020eternal}
            & $0.00_{0.00}$ & $12.39_{18.60}$ & $12.16_{18.60}$ & $1.22_{1.59}$ & $15.47_{16.50}$ & $0.10_{0.39}$ & $0.11_{0.20}$ & $22.27$\\
& LIP  \cite{foster2024information}
            & $0.00_{0.00}$ & $14.85_{22.29}$ & $14.58_{22.30}$ & $1.61_{2.10}$ & $15.19_{16.20}$ & $0.08_{0.31}$ & $0.10_{0.30}$ & $23.34$\\
& EMMN \cite{chundawat2023zero}
            & $0.30_{0.55}$ & $11.34_{17.02}$ & $10.24_{15.66}$ & $49.70_{64.98}$ & $46.86_{49.98}$ & $10.21_{39.53}$ & $36.32_{67.49}$ & $50.59$\\
& CLIP-LIP \cite{kravets2025zero}
            & $67.80_{100.00}$ & $68.44_{100.00}$ & $68.34_{100.00}$ & $76.78_{100.00}$ & $93.66_{99.90}$ & $25.91_{100.00}$ & $53.85_{100.00}$ & $82.08$\\
& TIFS \cite{zhang2025targeted}
            & $5.03_{9.30}$ & $61.61_{92.49}$ & $60.47_{92.50}$ & $65.17_{85.20}$ & $87.38_{93.21}$ & $21.70_{84.01}$ & $45.09_{83.79}$ & $88.84$\\
\cmidrule{2-10}
& \projectname \ (ours)
            & $1.20_{2.22}$ & $74.61_{100.00}$ & $67.27_{100.00}$ & $74.67_{97.62}$ & $89.70_{95.68}$ & $24.34_{94.23}$ & $52.04_{96.71}$ & $\textbf{97.43}$\\

\midrule
\multirow{8}{*}{RN101}
& Original & 70.60 & 75.39 & 74.91 & 81.16 & 96.46 & 29.16 & 55.25 & -- \\ 
\cmidrule{2-10}
& FT  \cite{warnecke2023machine}
            & $14.90_{21.10}$ & $63.84_{84.68}$ & $57.89_{77.28}$ & $1.70_{2.09}$ & $66.90_{69.35}$ & $1.87_{6.41}$ & $1.29_{2.33}$ & $45.86$\\
& GA  \cite{thudi2022unrolling}
            & $4.50_{6.37}$ & $25.86_{34.30}$ & $23.72_{31.66}$ & $3.47_{4.28}$ & $56.35_{58.39}$ & $1.97_{6.76}$ & $1.64_{2.97}$ & $33.14$\\
& Fisher \cite{golatkar2020eternal}
            & $0.00_{0.00}$ & $15.08_{20.00}$ & $14.98_{20.00}$ & $1.38_{1.70}$ & $17.65_{18.30}$ & $0.12_{0.41}$ & $0.17_{0.31}$ & $22.96$\\
& LIP  \cite{foster2024information}
            & $0.00_{0.00}$ & $16.13_{21.40}$ & $16.03_{21.40}$ & $2.27_{2.80}$ & $16.40_{17.00}$ & $0.09_{0.31}$ & $0.28_{0.51}$ & $23.34$\\
& EMMN \cite{chundawat2023zero}
            & $78.30_{100.00}$ & $6.47_{8.58}$ & $13.65_{18.22}$ & $45.29_{55.80}$ & $42.85_{44.42}$ & $10.71_{36.73}$ & $36.54_{66.14}$ & $31.28$\\
& CLIP-LIP \cite{kravets2025zero}
            & $3.60_{5.10}$ & $75.67_{100.00}$ & $68.50_{91.44}$ & $79.11_{97.47}$ & $90.88_{91.22}$ & $26.83_{92.01}$ & $55.25_{100.00}$ & $95.72$\\
& TIFS \cite{zhang2025targeted}
            & $8.97_{12.71}$ & $73.96_{98.10}$ & $73.49_{98.10}$ & $69.15_{85.20}$ & $91.54_{94.90}$ & $21.49_{83.98}$ & $36.52_{66.10}$ & $86.20$\\ 
\cmidrule{2-10}
& \projectname \ (ours)       
            & $0.00_{0.00}$ & $82.76_{100.00}$ & $74.48_{99.43}$ & $78.04_{96.16}$ & $92.85_{96.26}$ & $26.05_{89.33}$ & $53.11_{96.13}$ & $\textbf{96.76}$\\

\bottomrule
\end{tabular}}

\label{tab:cifar10_1}
\end{table*}

\begin{figure}[htbp]
    \centering
    \includegraphics[width=0.9\linewidth]{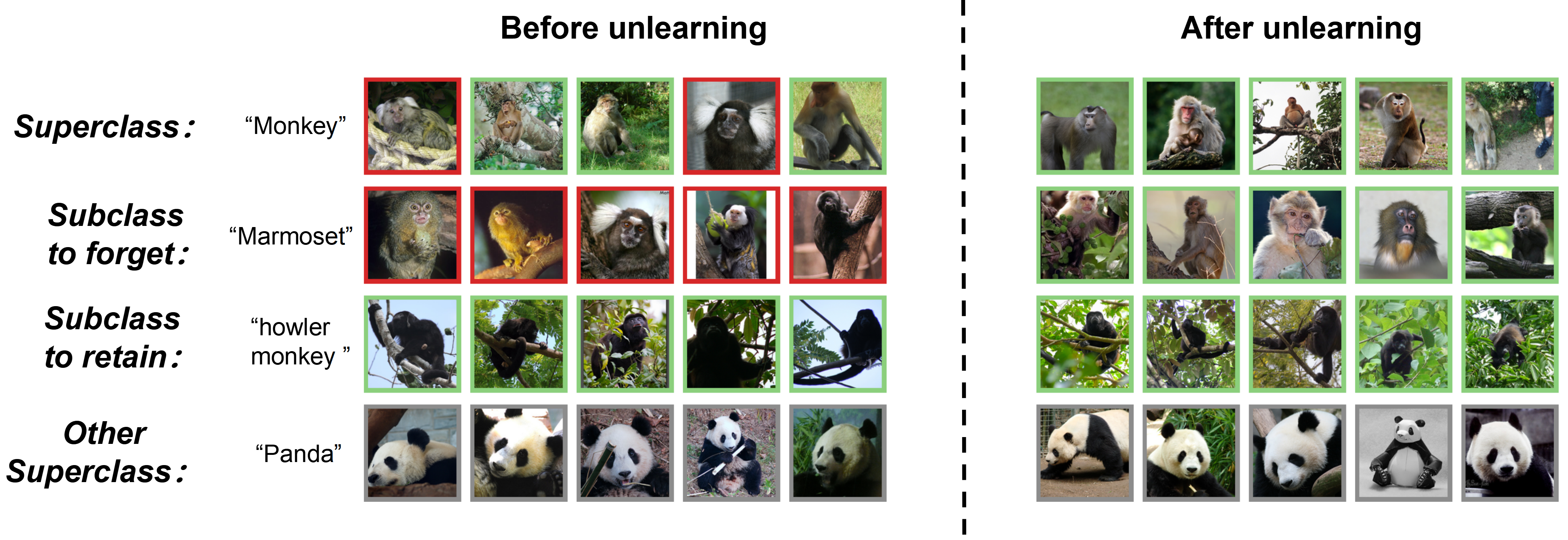}
    \caption{Retrieval visualization for in-domain forgetting on ImageNet-1K. The target subgroup \emph{marmoset} is highlighted in red, while other \emph{monkey} subgroups are highlighted in green. }
    \label{fig:retrieval_imagenet}
    \vspace{-1em}
\end{figure}

\partitle{Retrieval Visualization}
We further provide retrieval results before and after unlearning to qualitatively evaluate whether \projectname\ removes the target knowledge while preserving related non-target semantics. Fig.~\ref{fig:retrieval_imagenet} shows the in-domain case on ImageNet-1K, where the target subgroup \emph{marmoset} is removed from the superclass \emph{monkey}. Before unlearning, the retrieved results contain multiple target subgroup samples. After unlearning, these target samples are largely removed, while other monkey subgroups and unrelated superclasses such as \emph{panda} remain retrievable. This indicates that \projectname\ can suppress the target subgroup without collapsing the broader superclass semantics. Additional retrieval results on CIFAR-10 are provided in the supplementary material.

\begin{figure}[htbp]
    \centering
    \includegraphics[width=0.9\linewidth]{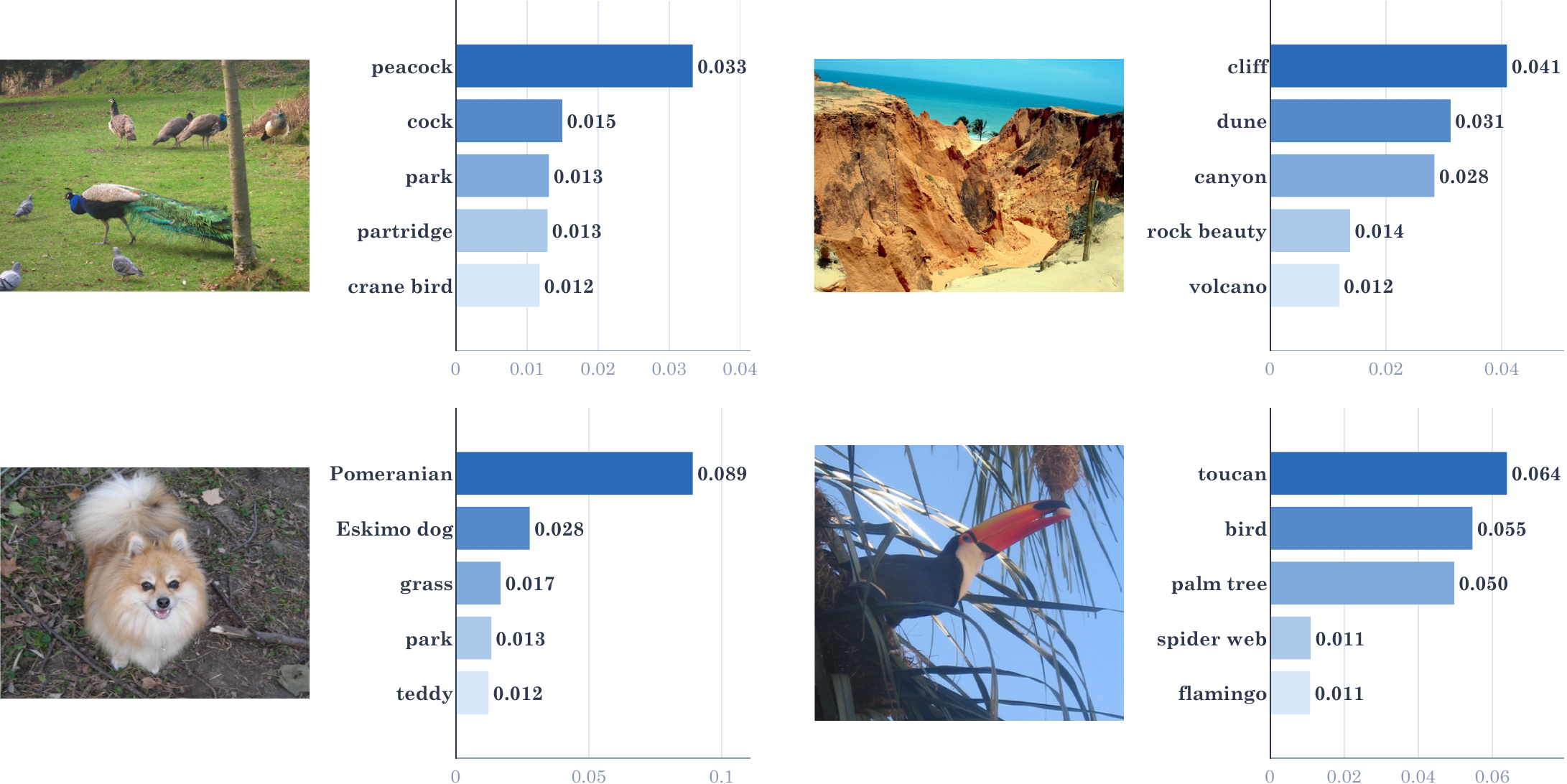}
    \caption{Visualization of the top-5 concepts obtained by \projectname.}
    \label{fig:visulization}
    \vspace{-1em}
\end{figure}

\partitle{Effectiveness of Concept Decomposition}
Fig.~\ref{fig:visulization} visualizes the top activated concepts produced by our sparse decomposition. The selected concepts are well aligned with the image content, covering both object-level semantics and contextual semantics. The decomposition is also sparse, with only a few concepts receiving dominant weights. These results suggest that our decomposition provides an interpretable and reliable concept-level interface for subsequent unlearning.

\subsection{Ablation Study}

\partitle{Impact of Each Loss Component}
Table~\ref{tab:ablation} shows that the three loss terms play complementary roles. $\mathcal{L}_{\text{forget}}$ is the main source of forgetting, as using it alone already drives the target accuracy close to zero. However, this comes at the cost of reduced retained and overall performance. Incorporating $\mathcal{L}_{\text{intra}}$ improves the preservation of non-target semantics within each instance, while $\mathcal{L}_{\text{global}}$ further stabilizes global knowledge and substantially boosts Avg. Score. The full objective achieves the best trade-off, yielding the highest retained accuracy, overall accuracy, and Avg. Score, which validates the necessity of all three components.

\begin{table}[htbp]
    \centering
    \caption{Ablation study of each loss component on CIFAR-10.}
    \resizebox{\columnwidth}{!}{
    \begin{tabular}{cccccccccccc}
    \toprule[1pt]
     \multirow{2}{*}{Config} & \multicolumn{3}{c}{Components} & \multicolumn{8}{c}{Metrics} \\ \cmidrule(r){2-4} \cmidrule(r){5-12}
     & $\mathcal{L}_{\text{forget}}$ & $\mathcal{L}_{\text{intra}}$ & $\mathcal{L}_{\text{global}}$ & Target$\downarrow$ & Retain$\uparrow$ & All$\uparrow$ & Food$\uparrow$ & STL$\uparrow$ & ObjectNet$\uparrow$ & ImageNet$\uparrow$ & Avg. Score$\uparrow$ \\ \midrule
     Original & - & - & - & $54.10$ & $66.61$ & $65.37$ & $76.49$ & $93.75$ & $25.83$ & $53.81$ & $-$ \\ \midrule
     1      & \checkmark & $\times$ & $\times$   & $0.00_{0.00}$ & $40.53_{60.85}$ & $36.48_{58.49}$ & $71.61_{93.62}$ & $80.40_{85.76}$ & $19.78_{76.58}$ & $41.80_{77.68}$ & $79.00$ \\
     2      & \checkmark & \checkmark & $\times$ & $0.00_{0.00}$ & $55.92_{83.95}$ & $50.33_{80.70}$ & $75.70_{98.97}$ & $85.71_{91.42}$ & $23.48_{90.90}$ & $49.60_{92.18}$ & $91.16$ \\
     3      & \checkmark & $\times$ & \checkmark & $0.00_{0.00}$ & $70.92_{100.00}$ & $63.83_{100.00}$ & $65.97_{86.25}$ & $81.11_{86.52}$ & $15.78_{61.09}$ & $41.81_{77.70}$ & $87.36$ \\ \midrule
     Ours   & \checkmark & \checkmark & \checkmark & $1.20_{2.22}$ & $74.61_{100.00}$ & $67.27_{100.00}$ & $74.67_{97.62}$ & $89.70_{95.68}$ & $24.34_{94.23}$ & $52.04_{96.71}$ & $\textbf{97.43}$ \\
    \bottomrule[1pt]
    \end{tabular}}
    \label{tab:ablation}
    \vspace{-1em}
\end{table}

\begin{figure}[htbp]
    \centering
    \subfloat[Vocabulary size]{
    \includegraphics[width=0.3\columnwidth]{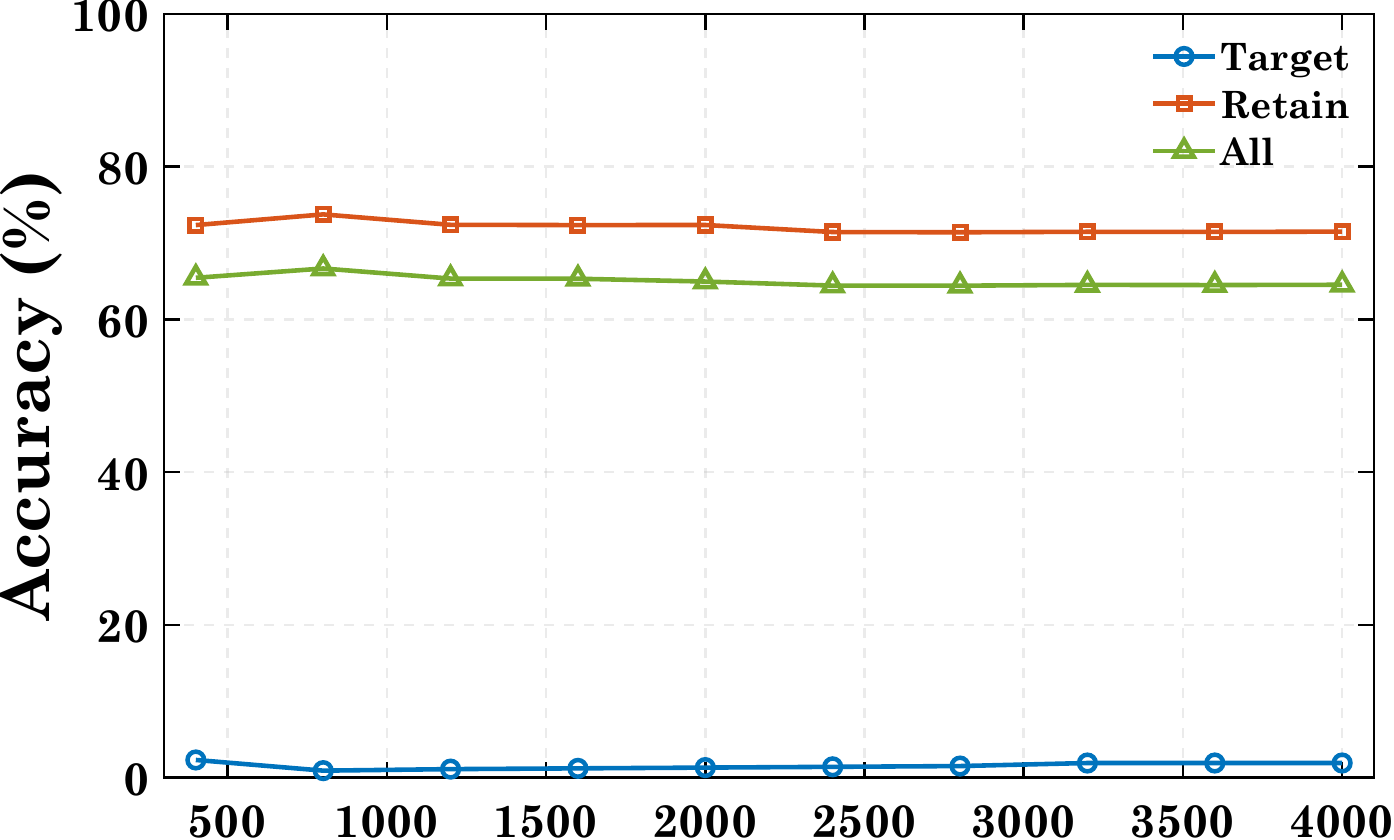}
    }
    \quad
    \subfloat[Sparsity regularization weight]{
    \includegraphics[width=0.3\columnwidth]{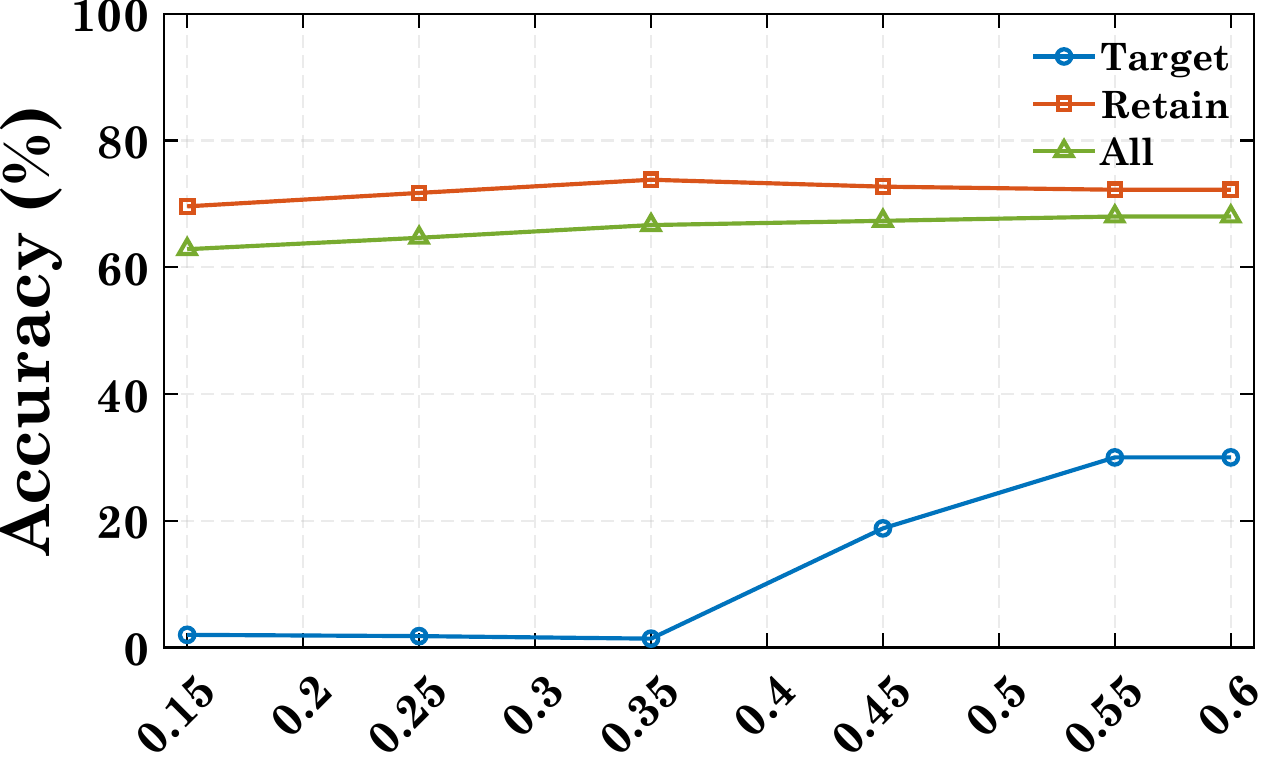}
    }
    \caption{Ablation study of the vocabulary size and sparsity regularization weight on CIFAR-10.}
    \label{fig:size and sparsity}
\end{figure}

\partitle{Impact of Vocabulary Size}
As shown in Fig.~\ref{fig:size and sparsity}(a), \projectname\ achieves strong performance even with a relatively small concept vocabulary. When the vocabulary size is set to 500, the target accuracy is already reduced to a very low level, while the retained and overall accuracy remain stable, indicating that a compact task-specific vocabulary is sufficient to capture the key semantics required for concept-level unlearning. This also suggests that \projectname\ does not rely on a large universal concept set, thereby reducing storage and decomposition overhead while avoiding interference from redundant concepts.

\partitle{Impact of and Sparsity Weight}
Fig.~\ref{fig:size and sparsity}(b) shows the influence of the sparsity regularization weight $\lambda_{\text{dec}}$. When $\lambda_{\text{dec}}$ is relatively small, the model achieves strong target forgetting while maintaining stable retained and overall performance. However, as $\lambda_{\text{dec}}$ becomes too large, the target accuracy increases noticeably, suggesting that excessive sparsity may weaken the representation of useful concepts and reduce the effectiveness of target concept suppression. 

\vspace{-1em}

\begin{figure}[htbp]
    \centering
    \subfloat[$\lambda_{\text{forget}}$]{
    \includegraphics[width=0.3\columnwidth]{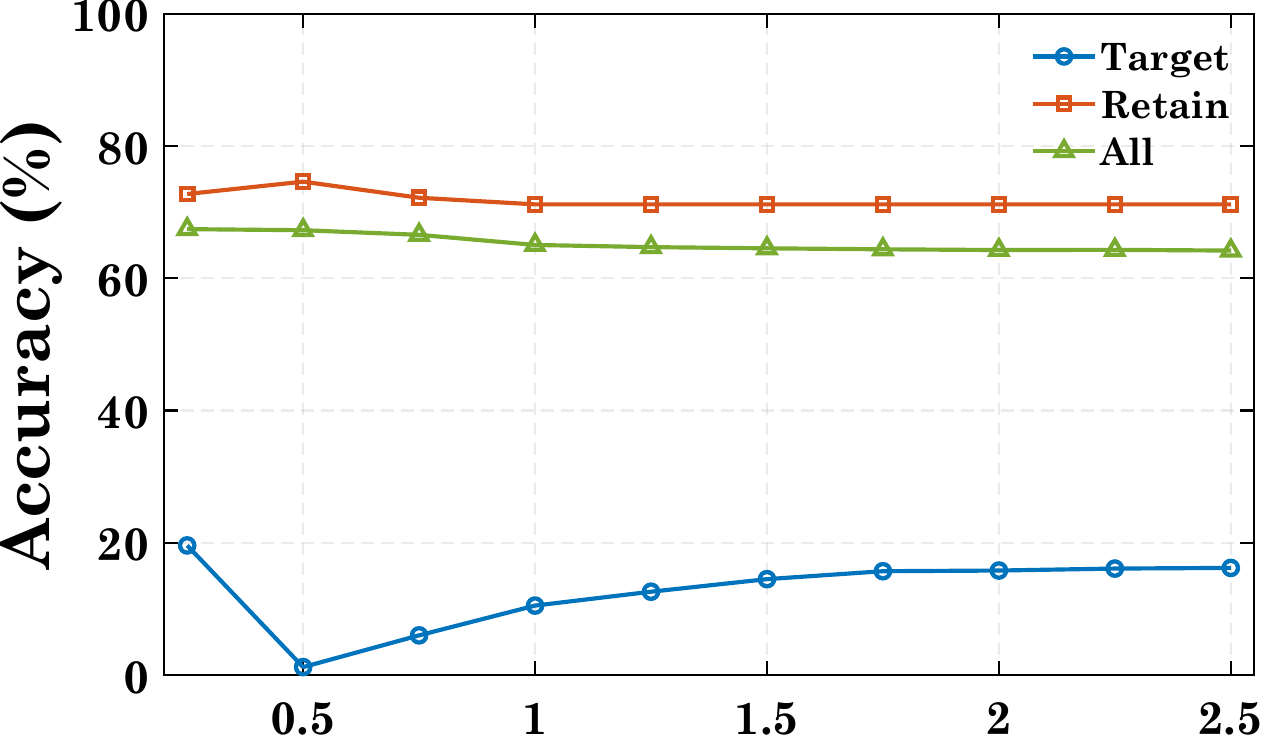}
    }
    \quad
    \subfloat[$\lambda_{\text{intra}}$]{
    \includegraphics[width=0.3\columnwidth]{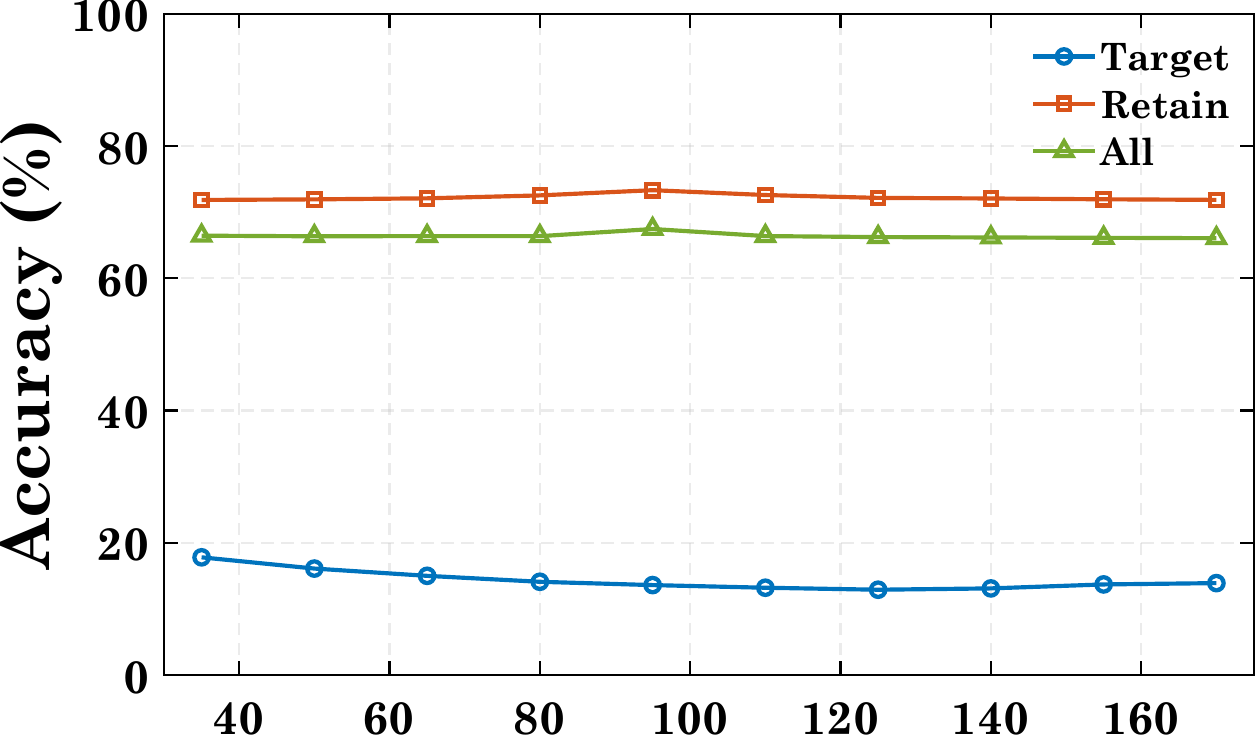}
    }
    \quad
    \subfloat[$\lambda_{\text{global}}$]{
    \includegraphics[width=0.3\columnwidth]{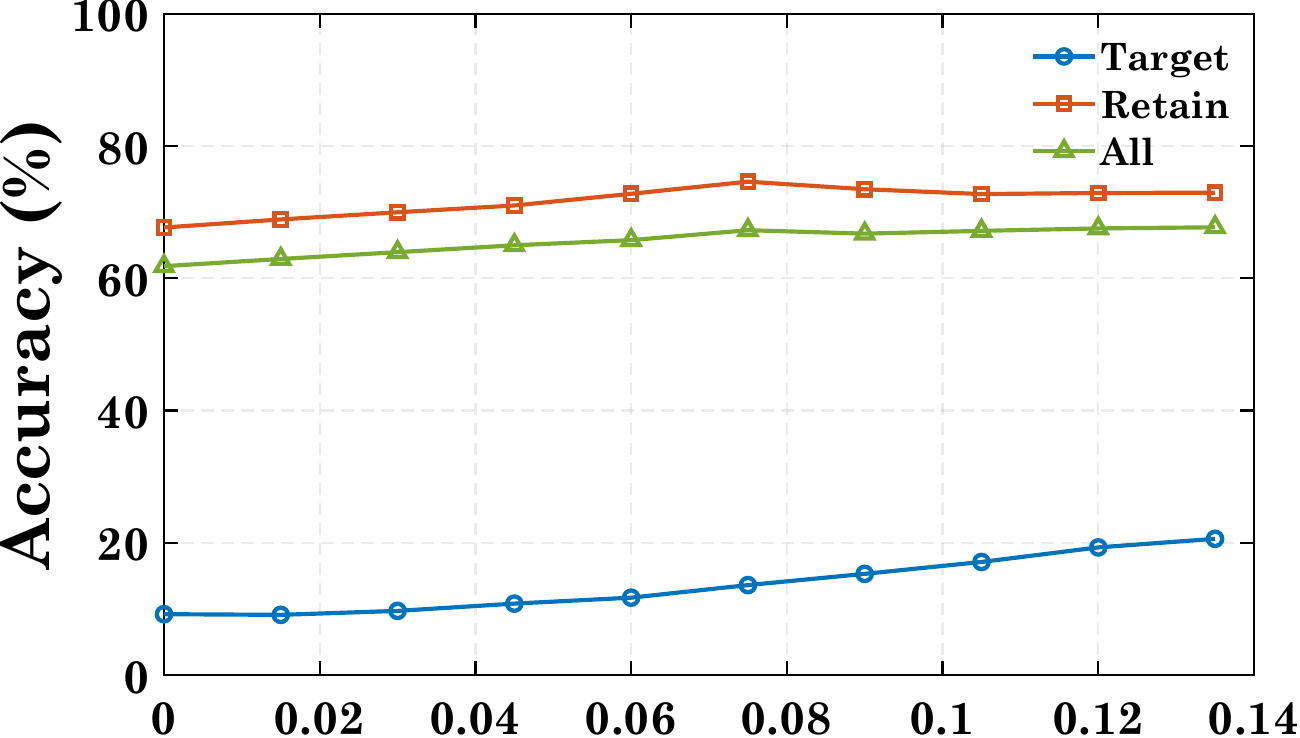}
    }
    \caption{Ablation study of the balancing hyperparameters on CIFAR-10.}
    \label{fig:hyperparameters}
    \vspace{-1em}
\end{figure}

\partitle{Impact of Balancing Hyperparameters}
We further analyze the influence of the balancing hyperparameters in Fig.~\ref{fig:hyperparameters}. For $\lambda_{\text{forget}}$, increasing its value generally strengthens target suppression, as reflected by the low forget accuracy. However, an overly large forgetting weight slightly increases the target accuracy and provides limited additional benefit to retained and overall performance. For $\lambda_{\text{intra}}$, the retained accuracy and overall accuracy remain stable across different values, suggesting that the intra-instance preservation term is not sensitive to the exact weighting and consistently helps maintain non-target semantics. For $\lambda_{\text{global}}$, increasing the weight improves retained and overall accuracy, but it also gradually increases the forget accuracy, indicating that a strong global preservation constraint may weaken target removal. Therefore, an appropriate balance among the three terms is necessary to achieve effective forgetting while preserving model utility.

\vspace{-1em}

\section{Conclusion}

In this paper, we propose \projectname, an interpretable concept-level machine unlearning framework for vision-language models. Instead of removing knowledge at the coarse instance level, \projectname\ decomposes visual representations into sparse combinations of task-specific semantic concepts and performs selective suppression on the target concepts. To preserve model utility, we further introduce intra-instance and global preservation objectives that maintain non-target semantics within each sample and retain general cross-modal knowledge on the remaining data. Extensive experiments under both in-domain and out-of-domain forgetting settings demonstrate that \projectname\ achieves an effective balance between target forgetting and model preservation.


\bibliographystyle{IEEEtran}
\bibliography{neurips_2026}

\appendix
\section{Additional Descriptions of \projectname}

Algorithm~\ref{alg:iced} summarizes the overall procedure of \projectname. The method consists of two stages. In the first stage, \projectname\ constructs a task-specific concept vocabulary from the forgetting set and performs sparse nonnegative decomposition to obtain interpretable concept weights for each forgetting sample. In the second stage, these concept weights are used to separate target and non-target semantic components, enabling concept-level unlearning with three complementary objectives: target concept suppression, intra-instance semantic preservation, and global cross-modal preservation. The first stage provides an interpretable concept-level representation of the forgetting data, while the second stage uses this representation to selectively remove target concepts and preserve non-target knowledge.

\begin{algorithm}[htbp]
\caption{Concept-level Machine Unlearning via \projectname}
\label{alg:iced}

\BlankLine
\textbf{Stage 1: Interpretable Concept Decomposition.}

Construct a task-specific concept vocabulary $\mathcal{C}_f$ from $\mathcal{D}_f$ using $\Phi$\;
Deduplicate and filter $\mathcal{C}_f$, and supplement common background concepts\;
Estimate modality means $\boldsymbol{\mu}_{\text{img}}$ and $\boldsymbol{\mu}_{\text{con}}$\;
Build the aligned concept dictionary $C$ using the CLIP text encoder $f_t$\;

\For{each $x \in \mathcal{D}_f$}{
    Compute the aligned image embedding $z$\;
    Solve sparse nonnegative decomposition:
    \[
    \boldsymbol{w}_x^\ast =
    \arg\min_{\boldsymbol{w}\in\mathbb{R}_{+}^{K}}
    \|C\boldsymbol{w}-z\|_2^2+\lambda_{\text{dec}}\|\boldsymbol{w}\|_1.
    \]
}

\BlankLine
\textbf{Stage 2: Concept-level Machine Unlearning.}

Freeze the text encoder $f_t$\;

\For{each training epoch}{
    \For{mini-batches $\mathcal{B}_f\subset\mathcal{D}_f$ and $\mathcal{B}_r\subset\mathcal{D}_r$}{
        Reconstruct target concept representations from $\{\boldsymbol{w}_x^\ast\}_{x\in\mathcal{B}_f}$\;
        Mask target concepts and reconstruct non-target representations\;
        Compute $\mathcal{L}_{\text{forget}}$, $\mathcal{L}_{\text{intra}}$, and $\mathcal{L}_{\text{global}}$\;
        Minimize
        \[
        \mathcal{L}_{\text{total}}
        =
        \lambda_{\text{forget}}\mathcal{L}_{\text{forget}}
        +
        \lambda_{\text{intra}}\mathcal{L}_{\text{intra}}
        +
        \lambda_{\text{global}}\mathcal{L}_{\text{global}}.
        \]
        Update only the image encoder $f_v$\;
    }
}
\end{algorithm}

\section{Proof}
\label{sec:proof}
\begin{proof}
By construction, the erased representation removes only the target component $C_Tw_T$ while keeping the non-target component $C_Rw_R$ and the residual $r$:
\begin{equation}
h(x)-\tilde h(x)=C_T\boldsymbol{w}_T.
\end{equation}
Therefore, for the target query $p_T$,
\begin{align}
\langle \boldsymbol{p}_T,h(x)\rangle-\langle \boldsymbol{p}_T,\tilde h(x)\rangle
&= \langle \boldsymbol{p}_T,C_T\boldsymbol{w}_T\rangle  \\
&= \sum_{\boldsymbol{c}_i\in C_T} \boldsymbol{w}_{T,i}\langle \boldsymbol{p}_T,\boldsymbol{c}_i\rangle .
\end{align}
Since $w_{T,i}\geq 0$ and $\langle p_T,c_i\rangle\geq \alpha$ for all target concepts, we obtain
\begin{equation}
\langle \boldsymbol{p}_T,h(x)\rangle-\langle \boldsymbol{p}_T,\tilde h(x)\rangle
\geq \alpha \sum_{\boldsymbol{c}_i\in C_T} \boldsymbol{w}_{T,i}
= \alpha \|\boldsymbol{w}_T\|_1 ,
\end{equation}
which proves Eq.~\eqref{eq:target_drop}.

Similarly, for any non-target query $p_R$,
\begin{align}
    |\langle \boldsymbol{p}_R,h(x)\rangle-\langle \boldsymbol{p}_R,\tilde h(x)\rangle|
    &= |\langle \boldsymbol{p}_R,C_T\boldsymbol{w}_T\rangle| \\
    &\leq \sum_{\boldsymbol{c}_i\in C_T} \boldsymbol{w}_{T,i} |\langle \boldsymbol{p}_R,\boldsymbol{c}_i\rangle| \\
    &\leq \eta \|\boldsymbol{w}_T\|_1 ,
\end{align}
which proves Eq.~\eqref{eq:retain_change}. Finally, the remaining target leakage
after erasure is bounded by
\begin{align}
    |\langle \boldsymbol{p}_T,\tilde h(x)\rangle|
    &= |\langle \boldsymbol{p}_T,C_R\boldsymbol{w}_R+\boldsymbol{r}\rangle| \\
    &\leq |\langle \boldsymbol{p}_T,C_R\boldsymbol{w}_R\rangle| + |\langle \boldsymbol{p}_T,\boldsymbol{r}\rangle| \\
    &\leq \sum_{\boldsymbol{c}_j\in \boldsymbol{C}_R} \boldsymbol{w}_{R,j}|\langle \boldsymbol{p}_T,\boldsymbol{c}_j\rangle|
        + \|\boldsymbol{p}_T\|_2\|\boldsymbol{r}\|_2 \\
    &\leq \beta\|\boldsymbol{w}_R\|_1+\epsilon_{\rm dec}.
\end{align}
This proves Eq.~\eqref{eq:target_leakage} and completes the proof.
\end{proof}

\section{Additional Implementation Details}
\label{sec:supp_baseline_details}
All experiments are conducted on a single NVIDIA RTX 4090 GPU with 48GB memory, using Python~3.10.19 and PyTorch~2.5.1. All CLIP backbones are initialized from the official OpenAI released pretrained weights\footnote{\url{https://github.com/openai/CLIP}}. For the implementation details of baselines, we follow the official implementations or commonly adopted settings from prior work as below.

\begin{itemize}
    \item \textbf{FT}\cite{warnecke2023machine} updates the image encoder using only the retain set $\mathcal{D}_r$, while keeping the text encoder frozen, and optimizes the standard CLIP contrastive loss. For ImageNet, we subsample approximately 50k retain samples to reduce computation, and train for 2 epochs using Adam with a learning rate of $1\times10^{-6}$ and a batch size of 128.
    \item \textbf{GA}\cite{thudi2022unrolling} performs gradient ascent on the current forget set $\mathcal{D}_u^{(t)}$ to maximize the CLIP loss and disrupt image--text alignment for target classes. Only the image encoder is updated, with gradient norm clipping applied for stability. We use Adam with a learning rate of $1\times10^{-6}$, a batch size of 128, and train for 2 epochs with gradient clipping set to 1.0.
    \item \textbf{Fisher}\cite{golatkar2020eternal} estimates the Fisher Information Matrix on the forgetting dataset and perturbs the model parameters by adding Gaussian noise whose variance is inversely proportional to the Fisher information, thereby reducing the model's sensitivity to the target data.
    \item \textbf{LIP}\cite{foster2024information} freezes the image encoder and updates the text projection matrix via a LoRA adapter, optimizing a combination of forget loss, retain loss, and a regularization term. To reduce computation on ImageNet, up to 512 retain classes are sampled. After optimization, the LoRA weights are merged back into the model. We use Adam with a learning rate of 0.01, LoRA rank 5, and run 2,000 iterations with $\lambda_1=0.3$, $\lambda_2$ dynamically adjusted, and $\lambda_3=1.0$.
    \item  \textbf{EMMN}\cite{chundawat2023zero} adopts a teacher--student framework with a learnable pseudo-image generator. The teacher model is fixed, while the student is updated using pseudo-images filtered by confidence, and training alternates between generator and student updates in a 1:5 ratio. We use Adam for both components, with a student learning rate of $1\times10^{-6}$, generator learning rate of $1\times10^{-5}$, batch size 128, temperature $T=1.0$, and confidence threshold $\delta=0.5$.
    \item \textbf{CLIP-LIP}\cite{kravets2025zero} extends LIP to the CLIP framework by applying low-rank adaptation to the text projection while preserving the pretrained image encoder. Similar to LIP, it updates the text projection using a low-rank LoRA module to suppress representations of the forgetting classes while maintaining alignment for retained classes. We follow the original implementation and use Adam with a learning rate of 0.01, LoRA rank 5, and 2,000 iterations.
    \item \textbf{TIFS}\cite{cai2025targeted} performs targeted forgetting by explicitly suppressing the model's responses on the forgetting dataset while preserving performance on retained data. Specifically, it optimizes the model to minimize prediction confidence on target classes while maintaining alignment on non-target samples. We follow the original implementation and update only the image encoder using Adam with a learning rate of $1\times10^{-6}$, batch size 128, and train for 2 epochs.
\end{itemize}

\section{Additional Experimental Results}

We provide additional comparison results under both in-domain and out-of-domain forgetting settings. These results complement the main paper by evaluating another forgetting target for each setting, using the same evaluation protocol and CLIP backbones.

\partitle{Additional comparison experiments in In-domain Forgetting}
Table~\ref{tab:imagenet_2} reports the results of \textit{Red Fox} forgetting on ImageNet-1K. Compared with existing baselines, \projectname\ achieves near-zero target accuracy while preserving stronger retained and overall ImageNet performance. In addition, \projectname\ maintains competitive transfer performance on Food, STL, ObjectNet, and CIFAR-10, leading to the best Avg. Score on both RN50 and RN101. These results further verify that \projectname\ can perform precise in-domain target removal with less degradation to non-target knowledge.

\begin{table*}[htbp]
\centering
\caption{Performance comparison with several baselines in \textit{Red Fox} forgetting on ImageNet-1K.}
\resizebox{\textwidth}{!}{
\begin{tabular}{cccccccccc}
\toprule
\multirow{2}{*}{Backbone} & \multirow{2}{*}{Method} & \multicolumn{3}{c}{ImageNet} & \multirow{2}{*}{Food$\uparrow$} & \multirow{2}{*}{STL$\uparrow$} & \multirow{2}{*}{ObjectNet$\uparrow$} & \multirow{2}{*}{CIFAR-10$\uparrow$} & \multirow{2}{*}{Avg. Score$\uparrow$} \\
\cmidrule{3-5}
& & Target$\downarrow$ & Retain$\uparrow$ & All$\uparrow$ & & & & & \\
\midrule

\multirow{8}{*}{RN50}
& Original
            & $48.00$ & $38.67$ & $51.94$ & $76.49$ & $93.75$ & $25.83$ & $68.84$ & $-$ \\
\cmidrule{2-10}
& FT  \cite{warnecke2023machine}
            & $6.00_{12.50}$ & $0.97_{2.51}$ & $20.96_{40.35}$ & $27.60_{36.08}$ & $54.33_{57.95}$ & $9.00_{34.84}$ & $14.14_{20.54}$ & $39.97$ \\
& GA  \cite{thudi2022unrolling}
            & $18.00_{37.50}$ & $36.18_{93.56}$ & $36.16_{69.62}$ & $55.04_{71.96}$ & $69.10_{73.71}$ & $15.86_{61.40}$ & $17.27_{25.09}$ & $65.40$ \\
& Fisher \cite{golatkar2020eternal}
            & $0.91_{1.90}$ & $1.08_{2.79}$ & $3.06_{5.89}$ & $0.54_{0.71}$ & $10.69_{11.40}$ & $0.59_{2.28}$ & $9.50_{13.80}$ & $19.29$ \\
& LIP  \cite{foster2024information}
            & $0.29_{0.60}$ & $1.86_{4.81}$ & $13.19_{25.39}$ & $7.65_{10.00}$ & $24.19_{25.80}$ & $4.24_{16.41}$ & $10.39_{15.09}$ & $28.13$ \\
& EMMN \cite{chundawat2023zero}
            & $22.00_{45.83}$ & $48.67_{100.00}$ & $31.04_{59.76}$ & $43.41_{56.75}$ & $59.42_{63.38}$ & $9.64_{37.32}$ & $10.70_{15.54}$ & $55.28$ \\
& CLIP-LIP \cite{kravets2025zero}
            & $33.50_{69.79}$ & $52.57_{100.00}$ & $53.72_{100.00}$ & $76.57_{100.00}$ & $93.83_{100.00}$ & $25.83_{100.00}$ & $68.55_{99.58}$ & $89.97$ \\
& TIFS \cite{zhang2025targeted}
            & $0.00_{0.00}$ & $38.17_{98.71}$ & $43.95_{84.62}$ & $63.43_{82.93}$ & $89.27_{95.22}$ & $16.16_{62.56}$ & $53.29_{77.41}$ & $85.92$ \\
\cmidrule{2-10}
& \projectname\ (ours)
            & $0.00_{0.00}$ & $49.99_{100.00}$ & $49.79_{95.86}$ & $74.72_{97.69}$ & $87.53_{93.37}$ & $23.37_{90.48}$ & $65.05_{94.49}$ & $\textbf{95.98}$ \\

\midrule
\multirow{8}{*}{RN101}
& Original
            & $52.00$ & $46.67$ & $54.33$ & $81.16$ & $96.46$ & $29.16$ & $73.82$ & $-$ \\
\cmidrule{2-10}
& FT  \cite{warnecke2023machine}
            & $18.00_{34.62}$ & $2.26_{4.84}$ & $23.05_{42.43}$ & $35.83_{44.15}$ & $54.30_{56.29}$ & $18.75_{64.30}$ & $16.41_{22.23}$ & $42.80$ \\
& GA  \cite{thudi2022unrolling}
            & $16.00_{30.77}$ & $38.97_{83.50}$ & $39.03_{71.84}$ & $62.59_{77.12}$ & $71.78_{74.41}$ & $18.75_{64.30}$ & $24.89_{33.72}$ & $67.73$ \\
& Fisher \cite{golatkar2020eternal}
            & $1.51_{2.90}$ & $1.45_{3.11}$ & $0.76_{1.40}$ & $0.41_{0.51}$ & $9.94_{10.30}$ & $0.26_{0.89}$ & $12.62_{17.10}$ & $18.63$ \\
& LIP  \cite{foster2024information}
            & $0.62_{1.19}$ & $1.96_{4.20}$ & $11.25_{20.71}$ & $11.93_{14.70}$ & $17.65_{18.30}$ & $6.18_{21.19}$ & $11.81_{16.00}$ & $27.70$ \\
& EMMN \cite{chundawat2023zero}
            & $52.00_{100.00}$ & $49.33_{100.00}$ & $24.90_{45.83}$ & $38.61_{47.57}$ & $66.79_{69.24}$ & $10.56_{36.21}$ & $14.17_{19.20}$ & $45.44$ \\
& CLIP-LIP \cite{kravets2025zero}
            & $0.50_{0.96}$ & $52.25_{100.00}$ & $49.88_{91.81}$ & $78.10_{96.23}$ & $92.95_{96.36}$ & $25.21_{86.45}$ & $67.40_{91.30}$ & $94.46$ \\
& TIFS \cite{zhang2025targeted}
            & $0.00_{0.00}$ & $48.49_{100.00}$ & $51.71_{95.18}$ & $79.04_{97.39}$ & $95.52_{99.03}$ & $19.47_{66.77}$ & $81.27_{100.00}$ & $94.05$ \\
\cmidrule{2-10}
& \projectname\ (ours)
            & $1.00_{1.92}$ & $54.98_{100.00}$ & $54.77_{100.00}$ & $78.68_{96.94}$ & $94.79_{98.27}$ & $26.93_{92.35}$ & $73.13_{99.07}$ & $\textbf{97.82}$ \\

\bottomrule
\end{tabular}}
\label{tab:imagenet_2}
\end{table*}

\partitle{Additional comparison experiments in Out-of-domain Forgetting}
Table~\ref{tab:cifar10_2} shows the results of \textit{Truck} forgetting on CIFAR-10. Existing methods often either fail to sufficiently remove the target class or suffer from severe utility degradation on retained and transfer datasets. In contrast, \projectname\ achieves effective target forgetting while preserving strong retained accuracy and overall model utility. The consistently higher Avg. Score across both backbones demonstrates the robustness of \projectname\ under out-of-domain forgetting.

\begin{table*}[htbp]
\centering
\caption{Performance comparison with several baselines in \textit{Truck} forgetting on CIFAR-10.}
\resizebox{\textwidth}{!}{
\begin{tabular}{cccccccccc}
\toprule
\multirow{2}{*}{Backbone} & \multirow{2}{*}{Method} & \multicolumn{3}{c}{CIFAR-10} & \multirow{2}{*}{Food$\uparrow$} & \multirow{2}{*}{STL$\uparrow$} & \multirow{2}{*}{ObjectNet$\uparrow$} & \multirow{2}{*}{ImageNet$\uparrow$} & \multirow{2}{*}{Avg. Score$\uparrow$} \\
\cmidrule{3-5}
& & Target$\downarrow$ & Retain$\uparrow$ & All$\uparrow$ & & & & & \\

\midrule

\multirow{9}{*}{RN50}
& Original
            & $84.80$ & $63.20$ & $65.36$ & $76.49$ & $93.75$ & $25.83$ & $53.81$ &$-$\\
\cmidrule{2-10}
& FT  \cite{warnecke2023machine}
            & $42.50_{50.12}$ & $62.17_{98.37}$ & $58.84_{90.02}$ & $1.50_{1.96}$ & $43.81_{46.73}$ & $0.82_{3.17}$ & $0.27_{0.50}$ & $41.52$\\
& GA  \cite{thudi2022unrolling}
            & $8.70_{10.26}$ & $31.83_{50.36}$ & $29.52_{45.17}$ & $1.24_{1.62}$ & $30.90_{32.96}$ & $0.67_{2.59}$ & $0.25_{0.46}$ & $31.84$\\
& Fisher \cite{golatkar2020eternal}
            & $0.00_{0.00}$ & $10.68_{16.90}$ & $11.05_{16.91}$ & $1.53_{2.00}$ & $19.69_{21.00}$ & $0.10_{0.39}$ & $0.11_{0.20}$ & $22.49$\\
& LIP  \cite{foster2024information}
            & $0.00_{0.00}$ & $11.94_{18.89}$ & $12.35_{18.90}$ & $1.15_{1.50}$ & $12.56_{13.40}$ & $0.15_{0.58}$ & $0.16_{0.30}$ & $21.94$\\
& EMMN \cite{chundawat2023zero}
            & $0.00_{0.00}$ & $11.12_{17.59}$ & $10.01_{15.31}$ & $51.01_{66.69}$ & $54.02_{57.62}$ & $10.49_{40.61}$ & $37.26_{69.24}$ & $52.44$\\
& CLIP-LIP \cite{kravets2025zero}
            & $77.70_{91.63}$ & $68.56_{100.00}$ & $68.68_{100.00}$ & $76.83_{100.00}$ & $93.92_{100.00}$ & $25.95_{100.00}$ & $53.85_{100.00}$ & $86.91$\\
& TIFS \cite{zhang2025targeted}
            & $1.44_{1.70}$ & $38.99_{61.69}$ & $40.33_{61.70}$ & $63.41_{82.90}$ & $75.66_{80.70}$ & $21.15_{81.88}$ & $46.87_{87.10}$ & $79.18$\\
\cmidrule{2-10}
& \projectname \ (ours)
            & $1.60_{1.89}$ & $70.63_{100.00}$ & $63.73_{97.51}$ & $75.15_{98.25}$ & $87.20_{93.01}$ & $23.14_{89.59}$ & $49.13_{91.30}$ & $\textbf{95.40}$\\

\midrule
\multirow{9}{*}{RN101}
& Original & $81.60$ & $74.17$ & $74.91$ & $81.16$ & $96.46$ & $29.16$ & $55.25$ & $-$\\
\cmidrule{2-10}
& FT  \cite{warnecke2023machine}
            & $33.10_{40.56}$ & $62.60_{84.40}$ & $58.32_{77.85}$ & $3.31_{4.08}$ & $65.70_{68.11}$ & $1.91_{6.55}$ & $1.40_{2.53}$ & $43.28$\\
& GA  \cite{thudi2022unrolling}
            & $7.30_{8.95}$ & $33.30_{44.90}$ & $30.70_{40.98}$ & $3.33_{4.10}$ & $60.44_{62.66}$ & $1.67_{5.73}$ & $1.21_{2.19}$ & $35.94$\\
& Fisher \cite{golatkar2020eternal}
            & $0.00_{0.00}$ & $12.24_{16.50}$ & $12.36_{16.50}$ & $1.38_{1.70}$ & $14.85_{15.40}$ & $0.12_{0.41}$ & $0.11_{0.20}$ & $21.53$\\
& LIP  \cite{foster2024information}
            & $0.00_{0.00}$ & $14.61_{19.70}$ & $14.76_{19.70}$ & $2.27_{2.80}$ & $16.40_{17.00}$ & $0.12_{0.41}$ & $0.17_{0.31}$ & $22.85$\\
& EMMN \cite{chundawat2023zero}
            & $54.00_{66.06}$ & $13.93_{18.78}$ & $12.54_{16.74}$ & $44.10_{54.34}$ & $41.95_{43.49}$ & $10.93_{37.48}$ & $35.55_{64.34}$ & $38.44$\\
& CLIP-LIP \cite{kravets2025zero}
            & $35.30_{43.26}$ & $72.22_{97.37}$ & $72.53_{96.82}$ & $79.05_{97.40}$ & $95.75_{99.26}$ & $26.90_{92.25}$ & $52.25_{94.57}$ & $90.63$\\
& TIFS \cite{zhang2025targeted}
            & $2.86_{3.50}$ & $65.57_{88.41}$ & $66.22_{88.40}$ & $65.41_{80.59}$ & $90.38_{93.70}$ & $24.55_{84.19}$ & $41.33_{74.81}$ & $86.66$\\
\cmidrule{2-10}
& \projectname \ (ours)
            & $0.00_{0.00}$ & $80.30_{100.00}$ & $72.28_{96.49}$ & $78.58_{96.82}$ & $89.46_{92.74}$ & $24.52_{84.09}$ & $48.41_{87.62}$ & $\textbf{93.97}$\\

\bottomrule
\end{tabular}}
\label{tab:cifar10_2}
\end{table*}

\begin{figure}[htbp]
    \centering
    \includegraphics[width=0.9\linewidth]{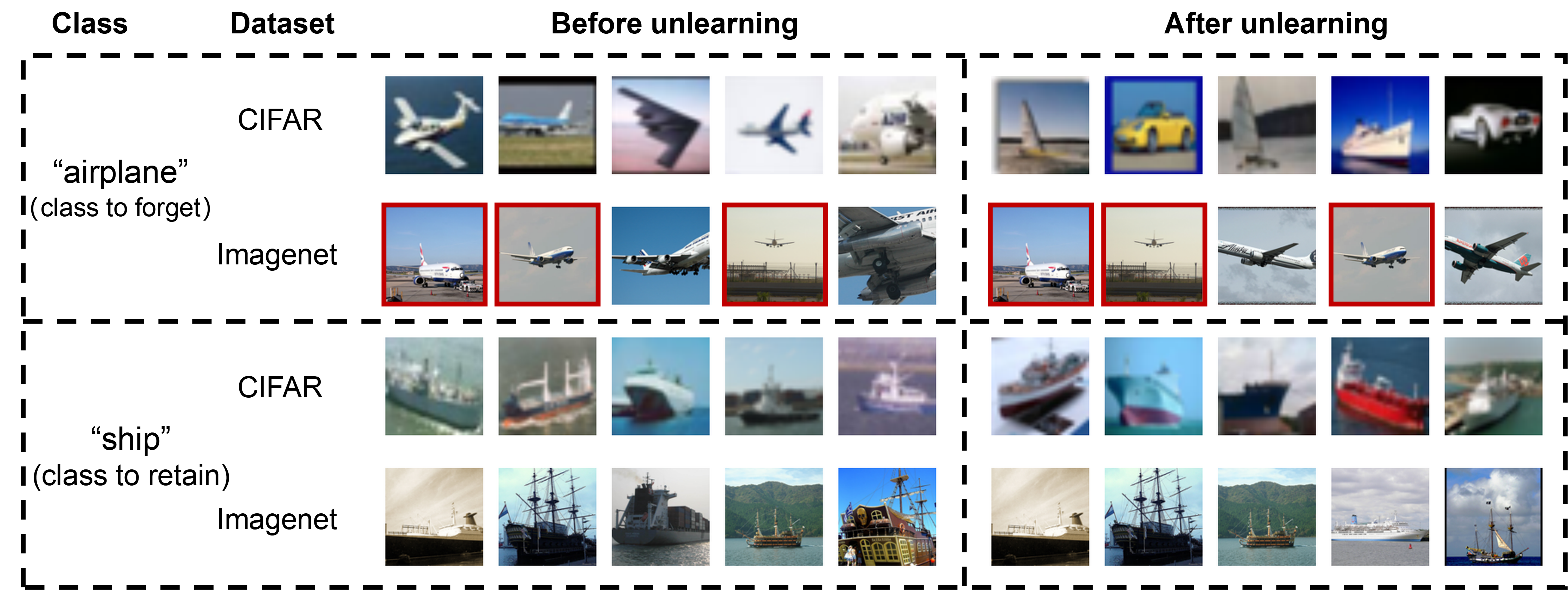}
    \caption{Retrieval visualization for out-of-domain forgetting on CIFAR-10. After forgetting the \emph{airplane} class, retrieval results for the retained \emph{ship} class remain stable, indicating preservation of non-target semantics.}
    \label{fig:retrieval_cifar}
\end{figure}

\partitle{Additional Retrieval Visualization}
Fig.~\ref{fig:retrieval_cifar} shows the out-of-domain case on CIFAR-10 with \emph{airplane} as the forgetting class and \emph{ship} as the retaining class. After unlearning, retrieval results for \emph{airplane} are significantly reduced or replaced, whereas the retrieved samples for \emph{ship} remain visually consistent before and after unlearning. These results suggest that \projectname\ performs targeted removal while preserving non-target visual-language alignment across datasets.

\section{Broader Impact and Limitations}

\subsection{Broader Impact}
\label{subsec:broader impact}
This work studies concept-level machine unlearning for vision-language models, aiming to remove target knowledge more precisely while preserving non-target semantics and general model utility. Compared with coarse image-level unlearning, the proposed framework provides a more interpretable interface for identifying and suppressing specific concepts, which may benefit applications involving privacy protection, data removal requests, harmful content mitigation, and controllable model editing. By reducing unnecessary degradation on unrelated knowledge, ICED also contributes to more reliable and transparent deployment of VLMs after unlearning.

At the same time, unlearning techniques may be misused to intentionally remove useful or socially important knowledge from models. Therefore, such methods should be applied with careful auditing, clear documentation of unlearning targets, and evaluation protocols that measure both forgetting effectiveness and unintended side effects.

\subsection{Limitations}
\label{subsec:limitations}
Although ICED enables more fine-grained unlearning, several limitations remain. First, the quality of concept decomposition depends on the constructed vocabulary. If the MLLM fails to extract important concepts or introduces noisy descriptions, the decomposition may be incomplete or less precise. Second, the proposed method relies on modality alignment and sparse decomposition, which approximate the visual representation with concept embeddings; this approximation may be imperfect when the target knowledge is highly abstract, relational, or not well captured by text concepts.

Third, our current experiments focus on CLIP-based models and classification-oriented evaluation settings. Extending the framework to larger generative VLMs, open-ended visual question answering, and more complex multimodal reasoning tasks remains an important direction. Finally, while ICED improves the trade-off between forgetting and preservation, it does not provide a formal guarantee that all traces of the target concept are removed. Future work may combine concept-level unlearning with stronger verification mechanisms and certified forgetting criteria.



\end{document}